\documentclass[runningheads]{llncs}
\usepackage[T1]{fontenc}
\usepackage{graphicx,verbatim}
\usepackage{booktabs}
\usepackage{multirow}  
\usepackage{stmaryrd} 
\usepackage{amssymb}
\usepackage{amsmath}
\usepackage{siunitx}
\usepackage[table]{xcolor}

\usepackage[colorlinks=true,
            linkcolor=blue,     
            citecolor=blue,     
            urlcolor=blue
           ]{hyperref}

\newcommand{\cf}{cf.~}
\newcommand{\venue}[1]{{\scriptsize\textcolor{gray}{[#1]}}}
\begin{document}
%
\title{Simple, Safe, and Overlooked: Reclaiming Sustainable Domain Generalization with Statistical Color Matching}
%
\titlerunning{Sustainable Domain Generalization with Statistical Color Matching}
%
\author{Sebastian Doerrich \and
Francesco Di Salvo \and
Shyam Nandan Rai \and
Marco Lents \and
Christian Ledig}
\authorrunning{S. Doerrich et al.}
%
\institute{xAILab Bamberg, University of Bamberg, Bamberg, Germany
\email{sebastian.doerrich@uni-bamberg.de}}
\maketitle              
\begin{abstract}
Hardware shifts, color variations, and changing patient characteristics between development and deployment routinely break trained medical image classifiers. Existing remedies fall short: standard color jittering provides insufficient diversity, while deep generative style transfer algorithms hallucinate features, destroy clinically relevant structures, and waste massive compute resources. To address this, we revisit classical statistical color matching and repurpose it as \emph{Colorist}, a highly efficient data augmentation strategy that applies global mean-standard deviation matching directly in the RGB color space. We demonstrate that this training-free, fully interpretable approach safely generates structurally intact domain variations, outperforming deep generative models in structural fidelity and color alignment. Across out-of-distribution histopathology, peripheral blood, dermatology, and retinal datasets, it improves balanced accuracy by up to +\qty{9}{\%} over state-of-the-art domain generalization regularizers and by +\qty{13}{\%} over an unaugmented baseline. Moreover, by avoiding neural networks in the augmentation loop, \emph{Colorist} preserves anatomical structure, minimizes carbon footprint, and integrates seamlessly into standard dataloaders. Together, these findings establish statistical matching as a safe, interpretable, yet overlooked alternative to deep architectures for clinical robustness. Source code is available at \url{https://github.com/sdoerrich97/colorist}.

\keywords{Domain Generalization \and Data Augmentation \and Color Transfer \and Medical Image Analysis \and Sustainable AI \and Interpretable AI}
\end{abstract}
\section{Introduction}
\label{sec:introduction}
\begin{figure}[htb]
\centering
\includegraphics[width=0.95\linewidth]{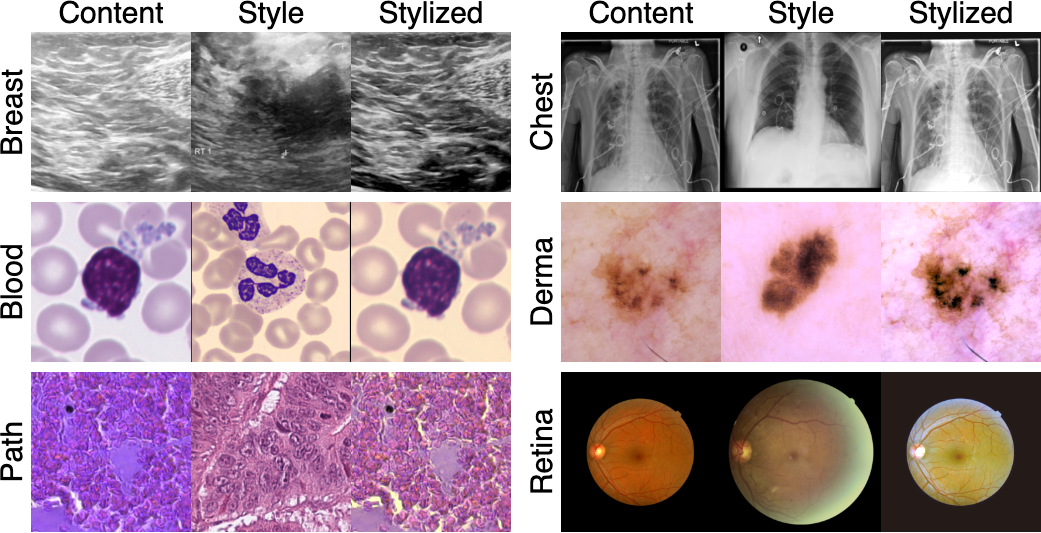}
\caption{Triplets of content, style, and stylized output images created by \emph{Colorist} across six distinct medical modalities demonstrate that simple global RGB matching safely transfers photometric shifts without corrupting clinical anatomy.}
\label{fig:PullFigure}
\end{figure}
Deploying deep learning-based decision support systems into unconstrained clinical environments requires models that withstand severe out-of-distribution (OOD) photometric shifts, such as variations in chemical staining and scanner calibration~\cite{Stacke2021}. Since the target distribution is unknown before deployment and may shift continuously, aligning the training and test distributions in advance is infeasible. Regularizing the training process to learn domain-invariant representations~\cite{Huang2020RSC,pezeshki2021SD,Ahuja2021InvariancePM,kim2021selfreg} offers only limited relief, as these techniques frequently fail to capture severe visual shifts. Data augmentation is a more practical alternative, yet existing strategies each introduce critical trade-offs. Traditional heuristics enforce orientation invariance but cannot simulate complex photometric variation, whereas automated search algorithms~\cite{Cubuk2019AutoAugmentLA,Cubuk2020RandAug,Muller2021ICCV,hendrycks2020augmix} optimize transformation policies that often apply semantically unsafe operations and corrupt diagnostically relevant features. Modality-specific augmentations, such as stain color augmentation for histopathology~\cite{TELLEZ2019101544}, address photometric shift directly but do not generalize across modalities and remain limited in scope. Deep generative methods, in contrast, separate anatomical content from style to synthesize diverse photometric variation~\cite{nguyen2024Contrimix,doerrich2024sgvits,doerrich2026stylizingvit}, but at a prohibitive training cost, a high carbon footprint, and a persistent risk of structural hallucination.

To capture fine-grained color patterns without these computational and structural burdens, we revisit traditional statistical color matching~\cite{reinhard2001color} and repurpose it as \emph{Colorist}, an overlooked yet highly effective training-time data augmentation strategy. Through a systematic evaluation of color spaces and matching algorithms, we show that global mean-standard deviation matching in the native RGB space performs competitively with decorrelated spaces, such as CIELAB or YCbCr, while remaining computationally cheaper. Because it relies exclusively on explicit mathematical equations rather than opaque black-box architectures, the resulting transformation is interpretable by design. This transparency makes \emph{Colorist} a safe alternative to compute-heavy generative models, preserving the spatial anatomical layout exactly and thereby avoiding the structural hallucinations these models can introduce during color transfer~(\figurename~\ref{fig:PullFigure}). Ultimately, \emph{Colorist} enables higher downstream classification accuracy, when used during training, than established augmentation protocols and representation-based domain generalization methods. We validate this across an extensive suite of 12 in-distribution and 7 covariate-shifted OOD datasets, spanning multiple modalities and dataset scales.
Our contributions can be summarized as follows:
\begin{itemize}
    \item We revisit statistical color matching and repurpose it as \emph{Colorist}, an efficient and interpretable training-time data augmentation strategy for robust out-of-distribution generalization across diverse clinical environments.
    \item We demonstrate that \emph{Colorist} outperforms established, compute-heavy deep generative models in both color transfer quality and structural preservation. As a per-pixel affine transform, it preserves the spatial anatomical layout exactly and thereby avoids the structural hallucinations of deep generative models, while providing a more sustainable solution through reduced computational overhead and a lower carbon footprint.
    \item We systematically evaluate multiple statistical color spaces and matching methods, proving that global mean-standard deviation matching in the native RGB space performs competitively with complex decorrelated spaces while avoiding dataloader conversion bottlenecks.
    \item We conduct extensive empirical validation across 12 in-distribution and 7 covariate-shifted OOD datasets, confirming that \emph{Colorist} surpasses, on average, established augmentation protocols and representation-based domain generalization methods across modalities and dataset sizes.
\end{itemize}
\section{Methodology}
\label{sec:method}
Clinically relevant photometric shift, whether from staining or scanner calibration, manifests primarily as a change in each channel's intensity distribution rather than in anatomy. This observation motivates statistical color matching: by aligning the first two moments of each color channel, we reproduce realistic photometric variation while leaving anatomical structure untouched, avoiding the structural risks and computational cost of deep generative models. Concretely, \emph{Colorist} transfers the per-channel color statistics of a style image $\mathbf{I}_s$ onto a content image $\mathbf{I}_c$, a design we formalize below and justify against more complex distribution-matching alternatives.

\subsubsection{Colorist} To provide a fast, structure-preserving foundation that enables on-the-fly application during model training, \emph{Colorist} treats the entire image as a single statistical region. The algorithm executes a global mean-variance matching independently for each native RGB color channel $C \in \{R, G, B\}$. For a pixel intensity $x_c$ of a specific channel in the content image $\mathbf{I}_c$, the algorithm computes the transferred pixel value $x'_c$ as:
\begin{equation}
    x'_c = (x_c - \mu_{c}) \left( \frac{\sigma_{s}}{\sigma_{c} + \epsilon_{var}} \right) + \mu_{s}
\label{eq:transfer}
\end{equation}
Here, $\mu$ and $\sigma$ denote the global mean and standard deviation of that specific channel for the respective images, and $\epsilon_{var} = 10^{-8}$ prevents division by zero. This simple linear transformation guarantees absolute preservation of the original anatomical layout. 

\subsubsection{Design Justification} Operating at this statistical level provides substantial advantages over related distribution matching techniques. Unlike standard histogram matching, which aligns cumulative distribution functions and frequently induces color quantization, contouring artifacts, or unnatural color shifts~\cite{Lefebvre2014}, \emph{Colorist} employs continuous linear scaling to preserve smooth, realistic gradients. Compared to Exact Histogram Matching (EHM)~\cite{Coltuc2006}, which relies on pixel-sorting algorithms with a complexity of $O(N \log N)$, \emph{Colorist} calculates its first and second moments in linear $O(N)$ complexity, maintaining high-throughput efficiency without compromising the training pipeline.
\section{Experiments and Results}
To assess the robustness of \emph{Colorist} and validate our core hypothesis that technical simplicity outcompetes generative complexity, we design a three-phase evaluation suite. First, we evaluate multiple color spaces and matching methods to justify the native RGB mean-standard deviation design. Second, we compare \emph{Colorist} against established style transfer methods to quantify structural fidelity and color alignment. Third, we evaluate downstream classification performance across an extensive array of in- and out-of-distribution (OOD) datasets.

To execute the initial color space evaluation and verify baseline anatomical integrity, we utilize the twelve 2D datasets from the MedMNIST+ collection~\cite{medmnistv2} (2--11 classes [C2--C11]; CC BY 4.0 / CC BY-NC 4.0), following their official data splits. 
The remaining datasets simulate severe clinical covariate shifts to test downstream robustness. To evaluate robustness against scanner and hospital variations, we employ the Camelyon17-WILDS benchmark~\cite{bandi2018detection,koh21a} ([C2]; CC0), training on hospitals H1 through H4 and testing on the unseen hospital H5. To assess robustness to staining protocol variations, we construct an Epithelium-Stroma benchmark ([C2]) that trains on H\&E-stained breast cancer images~\cite{Beck2011} (Public Domain) and tests exclusively on IHC-stained colorectal cancer images~\cite{Linder2012} (CC BY 4.0). For algorithmic fairness across demographic skin tones, we partition the Fitzpatrick17k~\cite{Groh2021} ([C3]; CC BY-NC-SA 3.0) and Diverse Dermatology Images (DDI)~\cite{ddidataset} ([C2]; Custom Research Use) datasets by Fitzpatrick Skin Type (FST I-II train, FST III-IV val, FST V-VI test). 
To measure generalization across microscopes and cell preparation techniques, we design two hematology benchmarks: a peripheral blood setup ([C13]; training on MLL23~\cite{ShetabBoushehri2025} and Acevedo20~\cite{acevedo2019recognition}, testing on Matek19~\cite{Matek2019}) and a bone marrow-to-peripheral blood shift task ([C13]; training on the BMC dataset~\cite{matek2021differentiation}, validation on Matek19, and testing on MLL23). All hematology datasets fall under CC BY 4.0. Finally, to evaluate robustness to fundus camera variations, we assemble a Retina dataset ([C5]) utilizing APTOS~\cite{aptos2019} (Non-Commercial Competition Use) and DeepDR~\cite{deepdr2022} (Permissive) for training, IDRiD~\cite{IDRiD} (CC-BY 4.0) for validation, and MESSIDOR-2~\cite{decenciere2014feedback} (Research Agreement) for testing. 
The chosen datasets range from 780 to 236,386 samples for MedMNIST+ and from approximately 650 to over 420,000 samples for the out-of-distribution clinical tasks. For standardization, we resize all images to $224 \times 224$ pixels via bilinear interpolation. 

\subsection{Validating the Efficiency of RGB Color Matching}
\label{sec:colorspace_eval}
To validate the architectural design of \emph{Colorist}, we conduct a systematic evaluation focusing on two primary axes: the selection of the matching algorithm and the impact of the underlying color space. Experiments are performed using the training sets of the twelve MedMNIST+ datasets. For each dataset and across three random seeds, we randomly sample 800 content and 800 style images to generate 800 stylized outputs per configuration. To quantify anatomical preservation, we compute SSIM and LPIPS~\cite{Zhang2018TheUE}. To measure the alignment of the color distributions, we evaluate FID~\cite{Heusel2017} and the Wasserstein Distance. Finally, we report ArtFID~\cite{Wright2022ArtFID} to assess anatomical retention and color alignment simultaneously. \tablename~\ref{tab:colorspace_eval} reports average performance across all twelve datasets.

\subsubsection{Results} Regarding the first axis, we justify the selection of the mean-standard deviation matching algorithm through a comparative analysis. Independent Friedman tests conducted across all eleven evaluated color spaces consistently reveal statistically significant differences in performance between the matching algorithms on the ArtFID metric (all $p < 0.01$). Subsequent post-hoc two-tailed Wilcoxon signed-rank tests further confirm that across all evaluated color spaces, mean-standard deviation matching yields statistically significant improvements over standard Histogram matching and EHM (\cf for RGB: $p=4.88 \times 10^{-4}$, $Z=-3.06$, $r=0.88$).
Regarding the second axis, we evaluate the impact of the underlying color space. The results show that RGB mean-standard deviation matching reports statistically equivalent results (Wilcoxon on ArtFID with a Bonferroni correction of $\alpha=0.0042$) to both YCbCr ($p=0.846$, $Z=-1.10$, $r=0.32$) and CIELAB ($p=0.042$, $Z=-2.04$, $r=0.59$), respectively.

\begin{table}[htb]
\centering
\caption{This quantitative evaluation analyzes structural fidelity and color alignment using mean-standard deviation matching across diverse color spaces. Results are reported as mean values across all twelve MedMNIST+ datasets.}
\label{tab:colorspace_eval}
\setlength{\tabcolsep}{9.5pt}
\begin{tabular}{l|cc|cc|c}
\toprule
\multirow{2.5}{*}{Color Space} & \multicolumn{2}{c}{Structure} & \multicolumn{2}{c}{Color} & \multicolumn{1}{c}{Fidelity} \\
\cmidrule(lr){2-3} \cmidrule(lr){4-5} \cmidrule(lr){6-6}
& LPIPS $\downarrow$ & SSIM $\uparrow$ & FID $\downarrow$ & W. Dist. $\downarrow$ & ArtFID $\downarrow$ \\
\midrule
HED & $0.30$ & $0.74$ & $55.21$ & $0.09$ & $76.27$ \\
HSI & $0.17$ & $0.82$ & $39.26$ & $0.05$ & $49.49$ \\
HSV & $0.16$ & $0.81$ & $33.51$ & $0.05$ & $42.48$ \\
CIELAB & $0.11$ & $0.84$ & $26.78$ & $0.04$ & $31.18$ \\
LCH & $0.15$ & $0.81$ & $31.18$ & $0.05$ & $38.16$ \\
LUV & $0.12$ & $0.83$ & $27.27$ & $0.04$ & $31.83$ \\
YCbCr & $0.11$ & $0.84$ & $26.30$ & $0.04$ & $30.61$ \\
YIQ & $0.11$ & $0.84$ & $32.44$ & $0.04$ & $37.29$ \\
YPbPr & $0.11$ & $0.84$ & $33.02$ & $0.04$ & $37.79$ \\
YUV & $0.11$ & $0.84$ & $28.54$ & $0.04$ & $33.00$ \\
\midrule
RGB & $0.11$ & $0.84$ & $26.32$ & $0.04$ & $30.66$ \\
\bottomrule
\end{tabular}
\end{table}
\subsection{Color Transfer and Structural Fidelity Evaluation}
We now evaluate how well \emph{Colorist} generalizes across our entire suite of in- and out-of-distribution clinical datasets. For this, we adopt the same evaluation protocol as before (800 image pairs, 3 random seeds) and benchmark \emph{Colorist} against thirteen complex deep generative models: \emph{Photorealistic Style Transfer} (Modflows~\cite{Larchenko2025AAAI}, WCT2~\cite{Yoo2019PhotorealisticST}), \emph{Artistic Style Transfer} (AdaIN~\cite{huang2017adain}, ArtFlow~\cite{artflow2021}, EFDM~\cite{Zhang2022ExactFD}, IEContrAST~\cite{chen2021iecontrast}, MAST~\cite{deng2020Mast}, SANET~\cite{Park2018ArbitraryST}, Styleformer~\cite{wu2021styleformer}, StyTr2~\cite{deng2021stytr2}), and \emph{Medical Style Transfer} (StylizingViT~\cite{doerrich2026stylizingvit}, ContriMix~\cite{nguyen2024Contrimix}, SGViTs~\cite{doerrich2024sgvits}). We train all artistic and medical style transfer methods from scratch on each dataset following their official guidelines and apply the general photorealistic methods out-of-the box without any fine-tuning.

\subsubsection{Results}
Table~\ref{tab:curated_ssim} reports the quantitative evaluation averaged across all datasets (MedMNIST+ and all OOD datasets). Artistic style transfer models consistently demonstrate poor structural retention, characterized by low SSIM and high LPIPS scores. Training-free photorealistic methods preserve the anatomy more effectively but still fall short of \emph{Colorist}, while specialized medical style transfer methods perform inconsistently across modalities or induce artifacts that degrade image quality. In contrast, \emph{Colorist} strictly preserves clinical geometry and achieves the highest structural preservation (SSIM $0.84$, LPIPS $0.12$) alongside superior color alignment (FID $28.33$, ArtFID $32.96$), validating its role as a safe foundation for photometric data augmentation.

\begin{table}[htb]
\centering
\caption{Quantitative evaluation of structural fidelity and color alignment averaged across all datasets. \emph{Colorist} successfully outperforms established deep learning architectures, achieving the best structural stability while transferring color statistics.}
\label{tab:curated_ssim}
\setlength{\tabcolsep}{6.5pt}
\begin{tabular}{l|cc|rc|r}
\toprule
\multirow{2.5}{*}{Method} & \multicolumn{2}{c}{Structure} & \multicolumn{2}{c}{Color} & \multicolumn{1}{c}{Fidelity} \\
\cmidrule(lr){2-3} \cmidrule(lr){4-5} \cmidrule(lr){6-6}
& LPIPS $\downarrow$ & SSIM $\uparrow$ & FID $\downarrow$ & W. Dist. $\downarrow$ & ArtFID $\downarrow$ \\
\midrule
AdaIN \venue{ICCV '17} & $0.33$ & $0.58$ & $70.29$ & $0.04$ & $97.93$ \\
ArtFlow \venue{CVPR '21} & $0.27$ & $0.60$ & $52.91$ & $0.03$ & $70.88$ \\
EFDM \venue{CVPR '22} & $0.35$ & $0.56$ & $77.76$ & $0.04$ & $109.27$ \\
IEContrAST \venue{NIPS '21} & $0.25$ & $0.70$ & $56.28$ & $0.08$ & $72.76$ \\
MAST \venue{ACM MM '20} & $0.27$ & $0.67$ & $55.04$ & $0.05$ & $73.27$ \\
SANET \venue{CVPR '19} & $0.25$ & $0.67$ & $46.86$ & $0.05$ & $62.36$ \\
Styleformer \venue{ICCV '21} & $0.54$ & $0.42$ & $139.48$ & $0.14$ & $222.70$ \\
StyTr2 \venue{CVPR '22} & $0.52$ & $0.60$ & $231.49$ & $0.19$ & $380.79$ \\
Contrimix \venue{COMPAY '24} & $0.40$ & $0.69$ & $115.18$ & $0.09$ & $168.70$ \\
SGViTs \venue{MICCAI '24} & $0.41$ & $0.67$ & $146.46$ & $0.08$ & $231.61$ \\
StylizingViT \venue{ISBI '26} & $0.15$ & $0.81$ & $80.47$ & $0.14$ & $112.24$ \\
Modflows \venue{AAAI '25} & $0.15$ & $0.83$ & $35.06$ & $0.04$ & $41.93$ \\
WCT2 \venue{ICCV '19} & $0.16$ & $0.81$ & $34.97$ & $0.04$ & $42.27$ \\
\midrule
\emph{Colorist} & $\mathbf{0.12}$ & $\mathbf{0.84}$ & $\mathbf{28.33}$ & $\mathbf{0.04}$ & $\mathbf{32.96}$ \\
\bottomrule
\end{tabular}
\end{table}

\begin{figure}[htb]
    \centering
    \includegraphics[width=\linewidth]{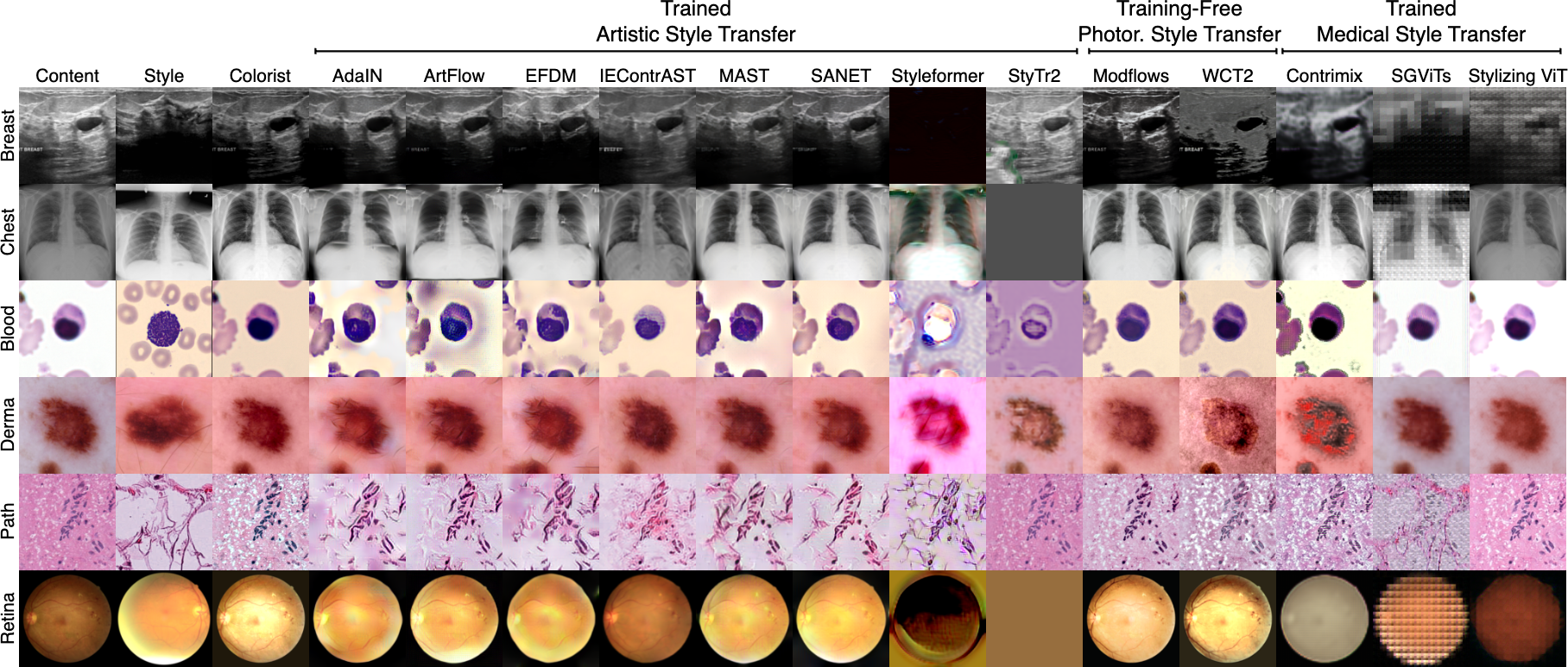}
    \caption{Qualitative comparison across six imaging modalities shows that complex style transfer corrupts clinical geometry while \emph{Colorist} maintains structural integrity.}
    \label{fig:qualitative_results}
\end{figure}

We visually confirm these findings through a qualitative comparison across six medical modalities (\figurename~\ref{fig:qualitative_results}). Deep generative methods consistently fail to provide safe photometric transformations while \emph{Colorist} maintains clinical geometry and successfully transfers the photometric style without injecting artificial structures.
\subsection{Downstream Classification Performance}
\label{sec:classification_results}
To establish the efficacy of \emph{Colorist} as a training-time augmentation, we evaluate downstream classification performance across our clinical benchmarks and report the balanced accuracy. We employ a DenseNet-121 classifier and compare \emph{Colorist} against an unaugmented baseline, traditional heuristic augmentations~\cite{hendrycks2020augmix,Cubuk2019AutoAugmentLA,Cubuk2020RandAug,zhong2020random,disalvo2024medmnistc,Muller2021ICCV}, and established feature-space regularizers~\cite{vapnik1998ERM,Ahuja2021InvariancePM,Huang2020RSC,pezeshki2021SD,kim2021selfreg} from DomainBed~\cite{gulrajani2021DomainBed}. All reference methods are applied using their default hyperparameters and protocols. We exclude domain-locked augmentations such as stain color normalization~\cite{TELLEZ2019101544}, whose modality-specific assumptions do not transfer across our benchmark. \emph{Colorist} is applied online with a 30\% probability, where for each source content image, we randomly sample a target style image directly from the training set. We train all methods independently on each dataset for 100 epochs across three seed runs using AdamW, cosine annealing (learning rate of 0.001), a batch size of 256, and early stopping (with 15 epochs).

\subsubsection{Results}
\emph{Colorist} attains the highest mean balanced accuracy (0.58) across the evaluated clinical shifts, outperforming both traditional augmentations and specialized domain generalization algorithms. As Table~\ref{tab:classificationBalAccuracy} details, standard geometric heuristics and basic color jittering fail to provide sufficient photometric diversity to bridge severe domain gaps, such as the staining variation in the Epithelium-Stroma (Epi-Str) benchmark. Similarly, complex DomainBed regularizers struggle to learn invariant representations, often performing barely above the unaugmented baseline despite their higher computational cost.
%
\begin{table}[htb]
\centering
\caption{Balanced Accuracy on the test set of each dataset, reported as the mean across three seed runs. Dataset abbreviations: MM (average across all MedMNIST datasets), C17 (Camelyon17), E-S (Epi-Str), Fitz (Fitzpatrick), Bld (Blood), Bne (Bone), Ret (Retina). \emph{Colorist} is bold when it ranks within the top three distinct values per column.
}
\label{tab:classificationBalAccuracy}
\setlength{\tabcolsep}{4.5pt}
\begin{tabular}{l|ccccccccc}
\toprule
Method & MM & C17 & E-S & Fitz & DDI & Bld & Bne & Ret & Avg \\
\midrule
Baseline & $0.79$ & $0.63$ & $0.61$ & $0.36$ & $0.48$ & $0.31$ & $0.16$ & $0.22$ & 0.45 \\
\midrule
AugMix~\cite{hendrycks2020augmix} & $0.79$ & $0.86$ & $0.51$ & $0.35$ & $0.50$ & $0.51$ & $0.30$ & $0.20$ & 0.50 \\
AutoAugment~\cite{Cubuk2019AutoAugmentLA} & $0.80$ & $0.70$ & $0.78$ & $0.40$ & $0.50$ & $0.52$ & $0.45$ & $0.20$ & 0.54 \\
Color Jitter & $0.78$ & $0.94$ & $0.71$ & $0.39$ & $0.50$ & $0.50$ & $0.36$ & $0.19$ & 0.55 \\
Gray Scale & $0.65$ & $0.74$ & $0.64$ & $0.34$ & $0.49$ & $0.10$ & $0.41$ & $0.24$ & 0.45 \\
RandAugment~\cite{Cubuk2020RandAug} & $0.81$ & $0.82$ & $0.63$ & $0.40$ & $0.49$ & $0.54$ & $0.51$ & $0.22$ & 0.55 \\
Random Erasing~\cite{zhong2020random} & $0.79$ & $0.68$ & $0.46$ & $0.36$ & $0.48$ & $0.26$ & $0.09$ & $0.23$ & 0.42 \\
Random Flip & $0.76$ & $0.69$ & $0.49$ & $0.40$ & $0.47$ & $0.29$ & $0.20$ & $0.23$ & 0.44 \\
Random Resized Crop & $0.79$ & $0.70$ & $0.61$ & $0.38$ & $0.48$ & $0.16$ & $0.13$ & $0.26$ & 0.44 \\
Targeted Augment~\cite{disalvo2024medmnistc} & $0.80$ & $0.83$ & $0.47$ & $0.40$ & $0.50$ & $0.50$ & $0.27$ & $0.20$ & 0.50 \\
TrivialAugment~\cite{Muller2021ICCV} & $0.81$ & $0.78$ & $0.69$ & $0.39$ & $0.50$ & $0.56$ & $0.44$ & $0.19$ & 0.55 \\
\midrule
ERM~\cite{vapnik1998ERM} & $0.79$ & $0.79$ & $0.65$ & $0.38$ & $0.50$ & $0.34$ & $0.14$ & $0.17$ & 0.47 \\
IB-ERM~\cite{Ahuja2021InvariancePM} & $0.79$ & $0.79$ & $0.66$ & $0.37$ & $0.50$ & $0.32$ & $0.21$ & $0.17$ & 0.48 \\
RSC~\cite{Huang2020RSC} & $0.78$ & $0.82$ & $0.80$ & $0.37$ & $0.50$ & $0.32$ & $0.12$ & $0.18$ & 0.49 \\
SD~\cite{pezeshki2021SD} & $0.80$ & $0.71$ & $0.59$ & $0.41$ & $0.50$ & $0.32$ & $0.15$ & $0.25$ & 0.47 \\
SelfReg~\cite{kim2021selfreg} & $0.78$ & $0.71$ & $0.69$ & $0.35$ & $0.49$ & $0.37$ & $0.19$ & $0.21$ & 0.47 \\
\midrule
\emph{Colorist} & $\mathbf{0.79}$ & $\mathbf{0.93}$ & $\mathbf{0.87}$ & $\mathbf{0.40}$ & $\mathbf{0.50}$ & $0.44$ & $\mathbf{0.46}$ & $0.21$ & $\mathbf{0.58}$ \\
\bottomrule
\end{tabular}
\end{table}
\section{Discussion and Conclusion}
\label{sec:discussion}


Our results show that the prevailing reliance on deep generative style transfer for medical domain generalization trades safety for capacity: such models risk structural hallucinations and incur substantial compute cost. \emph{Colorist} avoids this trade-off by transferring only per-channel color statistics in RGB space, preserving diagnostic anatomy by construction. Across nineteen datasets, this simple operation matches or surpasses both generative augmentations and feature-space regularizers while remaining training-free, interpretable, and sustainable.

\subsubsection{Limitations}
\emph{Colorist} models photometric shift through global color statistics and therefore does not address differences in acquisition geometry, spatial resolution, or anatomy. However, as a lightweight dataloader operation, it composes directly with the geometric and spatial augmentations that target these shifts.
On the Retina, Blood, and Bone benchmarks, no evaluated method reaches strong balanced accuracy, and \emph{Colorist} is no exception. The difficulty here is statistical rather than photometric: these tasks span up to thirteen classes and suffer pronounced class imbalance, leaving little for a color-based augmentation to exploit. Overcoming such regimes requires advances beyond training-time augmentation, such as imbalance-aware objectives or targeted sampling, which we leave to future work.

\begin{credits}
%
\subsubsection{\ackname} HPC resources were provided by the Erlangen National High Performance Computing Center (NHR@FAU) of Friedrich-Alexander-Universit\"at Erlangen-N\"urnberg (FAU). NHR@FAU hardware is partially funded by the German Research Foundation (DFG). This study was further funded through the Hightech Agenda Bayern (HTA) of the Free State of Bavaria, Germany.

\subsubsection{\discintname} The authors have no competing interests to declare that are relevant to the content of this article. 
\end{credits}
%
%
%
\bibliographystyle{splncs04}
\bibliography{mybibliography}

@article{Stacke2021,
   author = {Karin Stacke and others},
   issn = {21682208},
   issue = {2},
   journal = {IEEE Journal of Biomedical and Health Informatics},
   pages = {325-336},
   pmid = {33085623},
   publisher = {Institute of Electrical and Electronics Engineers Inc.},
   title = {Measuring Domain Shift for Deep Learning in Histopathology},
   volume = {25},
   year = {2021},
}

@article{TELLEZ2019101544,
title = {Quantifying the effects of data augmentation and stain color normalization in convolutional neural networks for computational pathology},
journal = {Medical Image Analysis},
volume = {58},
pages = {101544},
year = {2019},
issn = {1361-8415},
author = {David Tellez and others},
}

@article{gulrajani2021DomainBed,
    title={In Search of Lost Domain Generalization},
    author={Ishaan Gulrajani and David Lopez-Paz},
    journal={ICLR},
    year={2021},
}

@book{vapnik1998ERM,
  title={Statistical Learning Theory},
  author={Vapnik, Vladimir N},
  year={1998},
  publisher={Wiley-Interscience}
}

@article{Ahuja2021InvariancePM,
  title={Invariance Principle Meets Information Bottleneck for Out-of-Distribution Generalization},
  author={Kartik Ahuja and others},
  journal={NeurIPS},
  year={2021},
}

@article{Huang2020RSC,
    author="Huang, Zeyi and others",
    title="Self-challenging Improves Cross-Domain Generalization",
    journal="ECCV",
    year="2020",
    pages="124--140",
}

@article{pezeshki2021SD,
    title={Gradient Starvation: A Learning Proclivity in Neural Networks},
    author={Mohammad Pezeshki and others},
    journal={NeurIPS},
    editor={A. Beygelzimer and Y. Dauphin and P. Liang and J. Wortman Vaughan},
    year={2021},
}

@article{kim2021selfreg,
  title={Selfreg: Self-supervised contrastive regularization for domain generalization},
  author={Kim, Daehee and others},
  journal={ICCV},
  pages={9619--9628},
  year={2021}
}

@article{hendrycks2020augmix,
  title={{AugMix}: A Simple Data Processing Method to Improve Robustness and Uncertainty},
  author={Hendrycks, Dan and others},
  journal={ICLR},
  year={2020}
}

@article{Cubuk2019AutoAugmentLA,
  title={AutoAugment: Learning Augmentation Strategies From Data},
  author={Ekin Dogus Cubuk and others},
  journal={CVPR},
  year={2019},
  pages={113-123},
}

@article{Cubuk2020RandAug,
 author = {Cubuk, Ekin Dogus and others},
 journal = {NeurIPS},
 pages = {18613--18624},
 title = {RandAugment: Practical Automated Data Augmentation with a Reduced Search Space},
 volume = {33},
 year = {2020}
}

@article{zhong2020random,
title={Random Erasing Data Augmentation},
author={Zhong, Zhun and others},
journal={AAAI},
year={2020}
}

@misc{disalvo2024medmnistc,
      title={MedMNIST-C: Comprehensive benchmark and improved classifier robustness by simulating realistic image corruptions}, 
      author={Francesco Di Salvo and others},
      year={2024},
      eprint={2406.17536},
      archivePrefix={arXiv},
      primaryClass={eess.IV},
}

@article{Muller2021ICCV,
    author    = {M\"uller, Samuel G. and Hutter, Frank},
    title     = {TrivialAugment: Tuning-Free Yet State-of-the-Art Data Augmentation},
    journal = {ICCV},
    year      = {2021},
    pages     = {774-782}
}

@article{huang2017adain,
  title={Arbitrary Style Transfer in Real-time with Adaptive Instance Normalization},
  author={Huang, Xun and Belongie, Serge},
  journal={ICCV},
  year={2017}
}

@article{artflow2021,
 title={ArtFlow: Unbiased image style transfer via reversible neural flows},
 author={An, Jie and others},
 journal={CVPR},
 year={2021}
}

@article{Zhang2022ExactFD,
  title={Exact Feature Distribution Matching for Arbitrary Style Transfer and Domain Generalization},
  author={Yabin Zhang and others},
  journal={CVPR},
  year={2022},
  pages={8025-8035},
}

@inproceedings{Chen2021IEcontrast,
 author = {Chen, Haibo and others},
 booktitle = {NeurIPS},
 title = {Artistic Style Transfer with Internal-external Learning and Contrastive Learning},
 volume = {34},
 year = {2021}
}

@inproceedings{deng2020Mast,
  title={Arbitrary Style Transfer via Multi-Adaptation Network},
  author={Deng, Yingying and others},
  booktitle={Acm International Conference on Multimedia},
  year={2020},
 publisher = {ACM}
}

@article{Park2018ArbitraryST,
  title={Arbitrary Style Transfer With Style-Attentional Networks},
  author={Dae Young Park and Kwang Hee Lee},
  journal={CVPR},
  year={2018},
  pages={5873-5881},
}

@article{wu2021styleformer,
  title={StyleFormer: Real-Time Arbitrary Style Transfer via Parametric Style Composition},
  author={Wu, Xiaolei and others},
  journal={ICCV},
  pages={14618--14627},
  year={2021}
}

@article{deng2021stytr2,
      title={StyTr$^2$: Image Style Transfer with Transformers}, 
      author={Yingying Deng and others},
      journal={CVPR},
      year={2022}
}

@article{nguyen2024Contrimix,
  title = 	 {ContriMix: Scalable stain color augmentation for domain generalization without domain labels in digital pathology},
  author =       {Nguyen, Tan H. and others},
  journal = 	 {MICCAI Workshop on Computational Pathology},
  pages = 	 {121--130},
  year = 	 {2024},
  volume = 	 {254},
}

@article{doerrich2024sgvits,
    author={Sebastian Doerrich and others},
    title={Self-supervised Vision Transformer are Scalable Generative Models for Domain Generalization},
    journal={MICCAI},
    year={2024},
    pages={644--654},
    isbn={978-3-031-72117-5}
}

@misc{doerrich2026stylizingvit,
  title={Stylizing ViT: Anatomy-Preserving Instance Style Transfer for Domain Generalization},
  author={Sebastian Doerrich and others},
  year={2026},
  eprint={2601.17586},
  archivePrefix={arXiv},
  primaryClass={cs.CV}
}

@article{reinhard2001color,
  author    = {Reinhard, Erik and others},
  journal   = {IEEE Computer Graphics and Applications},
  title     = {Color transfer between images},
  year      = {2001},
  volume    = {21},
  number    = {5},
  pages     = {34--41},
}

@article{Yoo2019PhotorealisticST,
  title={Photorealistic Style Transfer via Wavelet Transforms},
  author={Jaejun Yoo and others},
  journal={ICCV},
  year={2019},
  pages={9035-9044},
}

@article{Larchenko2025AAAI,
title={Color Transfer with Modulated Flows},
volume={39},
number={4},
journal={AAAI},
author={Larchenko, Maria and others},
year={2025},
}

@ARTICLE{Coltuc2006,
  author={Coltuc, D. and others},
  journal={IEEE Transactions on Image Processing}, 
  title={Exact histogram specification}, 
  year={2006},
  volume={15},
  number={5},
  pages={1143-1152},
}

@inproceedings{Lefebvre2014,
    booktitle = {Eurographics 2014 - State of the Art Reports},
    title = {{A Survey of Color Mapping and its Applications}},
    author = {Faridul, H. S. and others},
    year = {2014},
    ISSN = {1017-4656},
}

@article{medmnistv2,
    title={MedMNIST v2-A large-scale lightweight benchmark for 2D and 3D biomedical image classification},
    author={Yang, Jiancheng and others},
    journal={Scientific Data},
    volume={10},
    number={1},
    pages={41},
    year={2023},
    publisher={Nature Publishing Group UK London}
}

@article{bandi2018detection,
  title={From detection of individual metastases to classification of lymph node status at the patient level: the CAMELYON17 challenge},
  author={Bandi, Peter and others},
  journal={IEEE Transactions on Medical Imaging},
  year={2018},
  publisher={IEEE}
}

@article{koh21a,
  title = 	 {WILDS: A Benchmark of in-the-Wild Distribution Shifts},
  author =       {Koh, Pang Wei and others},
  journal = 	 {ICML},
  pages = 	 {5637--5664},
  year = 	 {2021},
  volume = 	 {139},
}

@article{Beck2011,
    author = {Andrew H. Beck  and others},
    title = {Systematic Analysis of Breast Cancer Morphology Uncovers Stromal Features Associated with Survival},
    journal = {Science Translational Medicine},
    volume = {3},
    number = {108},
    year = {2011},
}

@article{Linder2012,
   author = {Nina Linder and others},
   issn = {17461596},
   issue = {1},
   journal = {Diagnostic Pathology},
   pages = {1-11},
   pmid = {22385523},
   publisher = {BioMed Central},
   title = {Identification of tumor epithelium and stroma in tissue microarrays using texture analysis},
   volume = {7},
   year = {2012},
}

@article{Groh2021,
  title={Evaluating Deep Neural Networks Trained on Clinical Images in Dermatology with the Fitzpatrick 17k Dataset},
  author={Matthew Groh and others},
  journal={CVPRW},
  year={2021},
  pages={1820-1828}
}

@article{ddidataset,
author = {Roxana Daneshjou and others},
title = {Disparities in dermatology AI performance on a diverse, curated clinical image set},
journal = {Science Advances},
volume = {8},
number = {32},
year = {2022},
}

@article{acevedo2019recognition,
  author  = {Acevedo, Andrea and others},
  title   = {Recognition of peripheral blood cell images using convolutional neural networks},
  journal = {Computer Methods and Programs in Biomedicine},
  volume  = {180},
  year    = {2019},
}

@Article{Matek2019,
author={Matek, Christian and others},
title={Human-level recognition of blast cells in acute myeloid leukaemia with convolutional neural networks},
journal={Nature Machine Intelligence},
year={2019},
day={01},
volume={1},
number={11},
pages={538-544},
}

@Article{ShetabBoushehri2025,
author={Shetab Boushehri, Sayedali and others},
title={A large expert-annotated single-cell peripheral blood dataset for hematological disease diagnostics},
journal={Scientific Data},
year={2025},
day={11},
volume={12},
number={1},
pages={1773},
}

@article{matek2021differentiation,
  author  = {Matek, Christian and others},
  title   = {Highly accurate differentiation of bone marrow cell morphologies using deep neural networks on a large image data set},
  journal = {Blood},
  volume  = {138},
  number  = {20},
  pages   = {1917--1927},
  year    = {2021},
}

@misc{aptos2019,
  author = {{Asia Pacific Tele-Ophthalmology Society}},
  title  = {{APTOS} 2019 Blindness Detection},
  year   = {2019},
  note   = {Kaggle Competition Dataset}
}

@article{deepdr2022,
  author  = {Liu, Ruiye and others},
  title   = {{DeepDRiD}: {D}iabetic {R}etinopathy {G}rading and {I}mage {Q}uality {E}stimation {C}hallenge},
  journal = {Patterns},
  volume  = {3},
  number  = {6},
  year    = {2022},
}

@Article{IDRiD,
AUTHOR = {Porwal, Prasanna and others},
TITLE = {Indian Diabetic Retinopathy Image Dataset (IDRiD): A Database for Diabetic Retinopathy Screening Research},
JOURNAL = {Data},
VOLUME = {3},
YEAR = {2018},
NUMBER = {3},
ARTICLE-NUMBER = {25},
ISSN = {2306-5729},
}

@article{decenciere2014feedback,
  author  = {Decenci{\`e}re, {\'E}tienne aand others},
  title   = {{FEEDBACK} {ON} {A} {PUBLICLY} {DISTRIBUTED} {IMAGE} {DATABASE}: {THE} {MESSIDOR} {DATABASE}},
  journal = {Image Analysis and Stereology},
  year    = {2014},
  volume  = {33},
  number  = {3},
  pages   = {231--234},
}

@article{Heusel2017,
 author = {Heusel, Martin and others},
 journal = {NeurIPS},
 title = {GANs Trained by a Two Time-Scale Update Rule Converge to a Local Nash Equilibrium},
 volume = {30},
 year = {2017}
}

@article{Zhang2018TheUE,
  title={The Unreasonable Effectiveness of Deep Features as a Perceptual Metric},
  author={Richard Zhang and others},
  journal={CVPR},
  year={2018},
  pages={586-595},
}

@inproceedings{Wright2022ArtFID,
    author="Wright, Matthias and Ommer, Bj{\"o}rn",
    title="ArtFID: Quantitative Evaluation of Neural Style Transfer",
    booktitle="Pattern Recognition",
    year="2022",
    pages="560--576",
}
%






\clearpage
\appendix
\setcounter{page}{1}\renewcommand{\thepage}{S\arabic{page}}

\renewcommand{\theHsection}{supp.\thesection}
\renewcommand{\theHtable}{supp.\thetable}
\renewcommand{\theHfigure}{supp.\thefigure}
\renewcommand{\theHequation}{supp.\theequation}

\numberwithin{table}{section}
\renewcommand{\thetable}{\thesection\arabic{table}}

\numberwithin{figure}{section}
\renewcommand{\thefigure}{\thesection\arabic{figure}}

\numberwithin{equation}{section}
\renewcommand{\theequation}{\thesection\arabic{equation}}


\begin{center}
    \Large\textbf{Supplementary Material:\\Simple, Safe, and Overlooked: Reclaiming Sustainable Domain Generalization with Statistical Color Matching}
\end{center}

\section{Overview of Supplementary Material}
\label{sec:overview_supp}
This supplement documents the experimental setup behind the main paper and reports the analyses behind claims the main text states without room to substantiate.
\textit{Section~\ref{sec:experimental_details_supp}} records the hardware, the containerized software environment, and the three random seeds behind every reported mean.
\textit{Section~\ref{sec:dataset_details_supp}} characterizes the nineteen-dataset evaluation suite, whose class counts range from two to fourteen and whose training splits reach an imbalance ratio of $717{:}1$, which is the statistical difficulty the main paper's limitations attribute the weak Retina, Blood, and Bone results to.
\textit{Section~\ref{sec:ablation_probability_supp}} shows that the $30\%$ application probability adopted in the main paper raises balanced accuracy on the covariate-shift datasets by $+0.148$ over no augmentation and that accuracy is then flat, while applying the transform to every batch carries a measurable in-distribution cost.
\textit{Section~\ref{sec:transform_properties_supp}} establishes that the transform preserves intensity rank order exactly within each channel, that clipping is its only non-affine step and touches $2.35\%$ of pixels, and that matching in the native RGB space costs nothing against decorrelated alternatives.
\textit{Section~\ref{sec:runtime_sustainability_supp}} reports that \emph{Colorist} runs on CPU at the latency of the fastest GPU-accelerated baselines and consumes less energy per sample than all thirteen of them.
\section{Experimental Details}
\label{sec:experimental_details_supp}
\subsection{Development}
All local development, prototyping, and initial model profiling were conducted on an isolated workstation equipped with a single NVIDIA RTX 6000 Ada GPU (48 GB VRAM), running Ubuntu 24.04.2 LTS and Python 3.13.

\subsection{Computation}
To execute our large-scale downstream classifier training and deep generative baseline benchmarking, jobs were deployed across the TinyGPU High-Performance Computing (HPC) cluster managed by NHR@FAU. We used NVIDIA A100 (40 GB) nodes for the compute-heavy pretraining and the resource-intensive fitting of all training-required baseline style transfer models. Downstream DenseNet-121 classifier training was parallelized across nodes using either 4$\times$ RTX 2080 Ti (11 GB), Tesla V100 (32 GB), or GeForce RTX 3080 (10 GB) configurations.

\subsection{Reproducibility}
Training jobs were scheduled onto whichever of the above node types was available, so the hardware is deliberately heterogeneous while the software is not. To hold dependencies, build isolation, and environment behavior fixed across the local workstation and the cluster, we used a standardized containerized environment, deployed locally as a production Docker image and translated into an Apptainer container on the cluster nodes. Reported differences are therefore between methods rather than between machines, since every method is trained and evaluated under the same container and the same protocol.
To mitigate variance from weight initialization and data-loader shuffling, all downstream classification and color transfer results report the arithmetic mean across three independent runs using the explicit random seeds $S \in \{71397589, 133560673, 265017005\}$. The per-image analyses of Section~\ref{sec:transform_properties_supp} and the runtime benchmark of Section~\ref{sec:runtime_sustainability_supp} are reported from one run at seed $265017005$, as neither depends on training stochasticity.
\section{Dataset Details}
\label{sec:dataset_details_supp}
We evaluate on nineteen datasets: the twelve in-distribution benchmarks of the MedMNIST+ collection and seven clinical datasets carrying severe out-of-distribution (OOD) covariate shifts. This is the suite every accuracy figure in the main paper is averaged over. Table~\ref{tab:dataset_stats_supp} reports, per dataset, the imaging modality, the native color channels, the number of target classes, the predefined splits, the training-split Imbalance Ratio ($\mathrm{IR}$), and the Normalized Shannon Entropy ($H$), the last two being the imbalance characteristics the main paper's limitations discuss. Both are reported because they capture different things: $\mathrm{IR}$ responds only to the extreme classes, while $H$ reflects the whole distribution. Bone Marrow carries the suite's most extreme ratio at $717{:}1$ yet a higher $H$ than DermaMNIST at $58.66$, because its imbalance is concentrated in the extreme classes rather than spread across the distribution. The MedMNIST+ entries and the OOD entries are independent cohorts even where a modality recurs: the Retina benchmark and RetinaMNIST, and the Peripheral Blood benchmark and BloodMNIST, are assembled from different source datasets rather than being the same data at two resolutions.

\begin{table}[htb]
\caption{Dataset details for the full evaluation suite. Columns give the clinical modality, the native color channels (Color; Gray denotes grayscale), the number of target classes (\# C), the predefined sample counts (Train / Val / Test), the Imbalance Ratio ($\mathrm{IR} = \max_c(n_c) / \min_c(n_c)$), and the Normalized Shannon Entropy ($H = -\frac{1}{\ln k}\sum_{c}P_c\ln P_c \in [0,1]$), with $H=1$ denoting a perfectly balanced training split. Both $\mathrm{IR}$ and $H$ are computed on the training split. Modality abbreviations: Histopath. (histopathology), BM (bone marrow), US (ultrasound), Photo. (photography), Micr. (microscopy). The upper block lists the seven out-of-distribution benchmarks and the lower block the twelve MedMNIST+ datasets. Imbalance varies by more than two orders of magnitude across the suite, from balanced (Camelyon17, $\mathrm{IR}=1.00$) to $717{:}1$ on Bone Marrow.}
\label{tab:dataset_stats_supp}
    \centering
    \setlength{\tabcolsep}{4pt}
    \resizebox{\textwidth}{!}{%
    \begin{tabular}{l l c r r r r}
        \toprule
        Dataset & Modality & Color & \# C & Train / Val / Test & IR & $H$ \\
        \midrule
        Camelyon17 & H\&E Histopath. & RGB & 2 & 302,436 / 34,904 / 85,054 & 1.00 & 1.0000 \\
        EpiStr & H\&E Histopath. & RGB & 2 & 8,337 / 5,920 / 1,376 & 1.42 & 0.9783 \\
        Peripheral Blood & Blood Micr. & RGB & 13 & 44,930 / 11,234 / 18,339 & 8.93 & 0.9270 \\
        Bone Marrow & BM Cytology & RGB & 13 & 147,799 / 18,365 / 41,621 & 717.50 & 0.8440 \\
        Fitzpatrick17k & Dermatoscopy & RGB & 3 & 7,755 / 6,089 / 2,168 & 4.88 & 0.7422 \\
        DDI & Dermatoscopy & RGB & 2 & 208 / 241 / 207 & 3.24 & 0.7876 \\
        Retina & Fundus Photo. & RGB & 5 & 4,862 / 455 / 1,744 & 5.22 & 0.8665 \\
        \midrule
        PathMNIST & Colon Pathology & RGB & 9 & 89,996 / 10,004 / 7,180 & 1.63 & 0.9943 \\
        ChestMNIST & Chest X-Ray & Gray & 14 & 78,468 / 11,219 / 22,433 & 96.62 & 0.8610 \\
        DermaMNIST & Dermatoscopy & RGB & 7 & 7,007 / 1,003 / 2,005 & 58.66 & 0.5808 \\
        OCTMNIST & Retinal OCT & Gray & 4 & 97,477 / 10,832 / 1,000 & 5.94 & 0.8361 \\
        PneumoniaMNIST & Chest X-Ray & Gray & 2 & 4,708 / 524 / 624 & 2.88 & 0.8235 \\
        RetinaMNIST & Fundus Photo. & RGB & 5 & 1,080 / 120 / 400 & 7.36 & 0.8744 \\
        BreastMNIST & Breast US & Gray & 2 & 546 / 78 / 156 & 2.71 & 0.8404 \\
        BloodMNIST & Blood Micr. & RGB & 8 & 11,959 / 1,712 / 3,421 & 2.74 & 0.9632 \\
        TissueMNIST & Kidney Cortex Micr. & Gray & 8 & 165,466 / 23,640 / 47,280 & 9.05 & 0.8675 \\
        OrganAMNIST & Abdominal CT & Gray & 11 & 34,561 / 6,491 / 17,778 & 4.54 & 0.9568 \\
        OrganCMNIST & Abdominal CT & Gray & 11 & 12,975 / 2,392 / 8,216 & 5.02 & 0.9496 \\
        OrganSMNIST & Abdominal CT & Gray & 11 & 13,932 / 2,452 / 8,827 & 5.64 & 0.9314 \\
        \bottomrule
    \end{tabular}}
\end{table}
\section{Ablation on Augmentation Probability}
\label{sec:ablation_probability_supp}
\begin{figure}[htb]
    \centering
    \includegraphics[width=\textwidth]{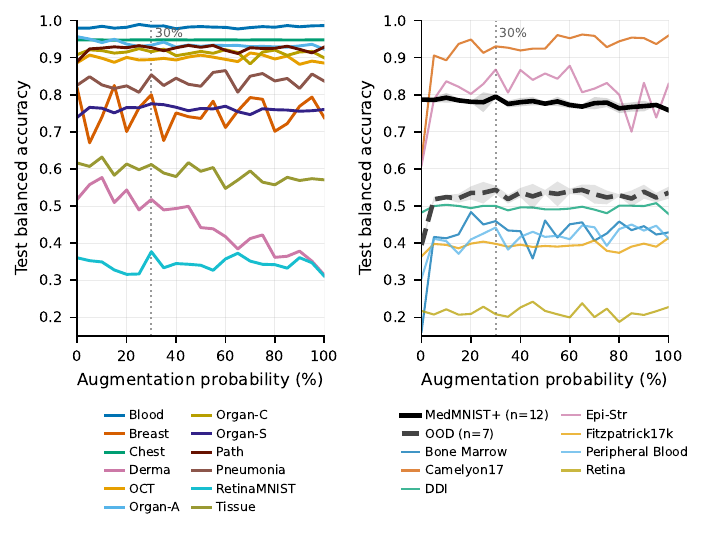}
    \caption{Downstream test balanced accuracy as a function of the online augmentation probability ($0\%$ to $100\%$). Left: the twelve MedMNIST+ datasets individually. Right: the two aggregates, MedMNIST+ ($n=12$, solid black) and the covariate-shift datasets (OOD, $n=7$, dashed), over the seven OOD datasets individually. Each aggregate is the unweighted mean over its group, computed within a seed and then averaged over the three seeds; the shaded band is the between-seed standard deviation. Every other coordinate is the mean over the same three seeds. The dotted vertical line marks the $30\%$ operating point adopted in the main paper. Both panels share the same vertical scale, so slopes are directly comparable across them. Both aggregates take their maximum at $30\%$, and only the in-distribution one declines appreciably under heavier application.}
    \label{fig:ablation_prob_supp}
\end{figure}

We sweep the sampling probability from $0\%$ (no augmentation) to $100\%$ (every batch stylized) in $5\%$ increments, training each configuration from scratch on all nineteen benchmarks across three seeds. The sweep serves two purposes: it locates the operating point the main manuscript adopts, and it separates the in-distribution cost of aggressive augmentation from its out-of-distribution benefit, which the aggregate figures of the main text cannot distinguish. Figure~\ref{fig:ablation_prob_supp} reports test balanced accuracy per dataset and for two aggregates. Each aggregate is the unweighted mean over its group, computed within a seed and then averaged across the three seeds, so the quoted spread is variation between seeds rather than between datasets.

\subsubsection{Results}
Both aggregates take their maximum at the $30\%$ probability adopted in the main manuscript, though in both cases the optimum is broad rather than sharp. Under covariate shift the aggregate rises from $0.3953$ without augmentation to $0.5438 \pm 0.0107$ at $30\%$, a gain of $+0.148$ balanced accuracy, and is then flat: the decline from that maximum to full application is $0.008$, which is $0.74\times$ the between-seed standard deviation at the operating point. Once the transform is applied at all, the covariate-shift result is therefore insensitive to how often.

In distribution the trade-off is real, and we do not claim the same insensitivity there. The MedMNIST+ aggregate gains only $+0.008$ over its unaugmented baseline of $0.787$, reaching $0.795 \pm 0.005$ at $30\%$, and falls to $0.758$ at full application. That decline of $0.037$ is $7.59\times$ the between-seed standard deviation at the operating point, it puts the augmented model below the unaugmented one, and at full application eight of the twelve MedMNIST+ datasets lose more than $0.02$ balanced accuracy from their own maximum. We therefore read $30\%$ as the point that buys the out-of-distribution gain at negligible in-distribution cost rather than as a free parameter. That maximum is not sharply separated from its neighbours, since the in-distribution aggregate reads $0.795 \pm 0.005$ at $30\%$ against $0.792 \pm 0.013$ at $10\%$, so the evidence supports a broad optimum near $30\%$ rather than that exact value.

\section{Properties of the Transform}
\label{sec:transform_properties_supp}
Because \emph{Colorist} is an explicit closed-form operation rather than a learned one, its behavior can be stated exactly rather than characterized empirically. This section makes three properties precise: what the transform guarantees about anatomical structure, where that guarantee stops holding, and why operating in the native, correlated RGB space is not the liability it is commonly assumed to be. All measurements below use 400 image pairs, and therefore $1{,}200$ image-channel instances, from the PathMNIST training split under seed 265017005.

\subsubsection{Results}
Per channel, the transform is the affine map $x \mapsto \alpha x + \beta$ with $\alpha = \sigma_{\mathrm{ref}} / (\sigma_{\mathrm{src}} + \epsilon_{var})$ and $\beta = \mu_{\mathrm{ref}} - \alpha \mu_{\mathrm{src}}$, matching Equation~(1) of the main paper. The stabilizer $\epsilon_{var} = 10^{-8}$ also fixes the degenerate case: a constant channel has $\sigma_{\mathrm{src}} = 0$ and is mapped to a constant rather than being undefined. Because the map is affine with $\alpha > 0$, it is exactly linear in the pixel intensities, so the Pearson correlation between source and output is $1.000000$ (minimum over all $1{,}200$ channel instances: $1.000000$), and being strictly increasing it also preserves the rank order of intensities. No spatial structure can be created or destroyed. Deep generative style transfer offers no comparable guarantee, which is what the structural gap in Table~2 of the main paper measures.

The single exception is clipping to the valid intensity range, the only non-affine step in the pipeline. After clipping, the mean Pearson correlation between source and output falls to $0.9916$ and $2.35\%$ of pixels are affected on average, while $38.2\%$ of channel instances are unaffected entirely. Structural loss in \emph{Colorist} therefore comes from saturation alone. A channel instance with no clipped pixels retains a correlation of exactly $1$, which is the case for $38.2\%$ of them, and the mean of $0.9916$ is reached at a mean clipped fraction of $2.35\%$. Because that fraction is computed from the image itself, the failure mode is auditable per image rather than opaque.

Reinhard et al.~\cite{reinhard2001color} apply the same mean-standard deviation matching in the decorrelated $l\alpha\beta$ space, on the reasoning that matching correlated channels independently distorts color. Our color space study (Table~\ref{tab:colorspace_eval} in the main paper) does not evaluate $l\alpha\beta$ itself, but it does evaluate four decorrelated spaces, CIELAB, LUV, YCbCr, and YUV, and RGB is statistically equivalent on ArtFID to the best of them, YCbCr, under the Bonferroni-corrected Wilcoxon test reported in the main paper. The penalty the decorrelation argument predicts therefore does not appear for this task. The correlation structure shows why. Pooling all three channels into a single vector, the source-output correlation is $0.903$, appreciably below $1$, but this reflects only the three channels receiving three different affine maps, which shifts the channels relative to one another; within any single channel the correlation remains exactly $1$. Since anatomical structure is carried within channels and not by their relative offsets, matching in RGB leaves the structure the classifier depends on intact while avoiding a forward and inverse color space conversion in the dataloader.

\section{Runtime, Throughput, and Energy Sustainability}
\label{sec:runtime_sustainability_supp}
A central argument for our framework is that it removes the energy cost of photometric augmentation without paying for it in data-loading throughput. To quantify this, we profile per-image latency (mean, median, and tail percentiles), throughput, and energy consumption measured with \texttt{codecarbon}, for \emph{Colorist} and for all thirteen deep generative baselines of Table~\ref{tab:curated_ssim} in the main paper, on GPU and on CPU. Each measurement processes one image per iteration, so an iteration is a sample and per-sample energy is directly comparable across rows. Every run discards $50$ warmup iterations before timing. \texttt{codecarbon} reads whole-device power, so a figure includes host draw alongside the accelerator and each run is measured on an otherwise idle machine.

Two properties of the comparison should be stated plainly rather than folded into a single number. First, \emph{Colorist} has no GPU implementation: it is per-channel arithmetic executed in the DataLoader worker processes, so a GPU would sit idle during it. Table~\ref{tab:runtime_sustainability_supp} therefore reports the baselines on both devices and \emph{Colorist} on CPU alone, which supports two readings: against the GPU block, a CPU-only statistical operation against GPU-accelerated networks, which is how each is actually deployed; and against the CPU block, a like-for-like comparison on a single device. Second, each baseline declares its own native input resolution and the harness honors it, so the resolutions are not common across rows: eight baselines run at $256 \times 256$, three at $224 \times 224$, and two at $512 \times 512$. Forcing every method to a single resolution would run it off the size it was designed for and make the timings meaningless, so the table reports the resolution per row and the comparisons below are grouped by it. \emph{Colorist} is measured at $256 \times 256$, the most common baseline resolution, rather than the $224 \times 224$ used for classifier training in the main paper.

Iteration counts differ by device: $10{,}000$ iterations on GPU and $1{,}000$ on CPU. At $10{,}000$ iterations the CPU runs alone would take roughly $134$ hours, $92\%$ of it in three methods (Styleformer at $21.2$ s per iteration, Modflows at $14.4$ s, and WCT2 at $8.7$ s). The count governs the precision of the estimate rather than its expected value, so means and medians remain comparable across devices and only the tail percentiles are looser on CPU.

\subsubsection{Results}
On the device each baseline is deployed on, avoiding neural operations leaves \emph{Colorist} on par with the fastest of them while it runs on CPU alone. At $2.53$ ms per sample it is slower than three of the thirteen GPU rows, within a factor of $1.7$ of the quickest measured (Contrimix at $1.50$ ms, on fewer pixels), and faster than the remaining ten by factors of $1.25$ to $57$.

On a matched device the comparison is no longer close. Against the same thirteen methods measured on CPU, \emph{Colorist} is faster than every one of them, by $2.5\times$ (Contrimix) to $8366\times$ (Styleformer, at $21.2$ s per image), and consumes between $5.4\times$ and $10{,}601\times$ less energy per sample. This is the reading that isolates the cost of the augmentation itself from the cost of the hardware it is given.

Energy separates the methods at every resolution, and it does so against the GPU rows as well. Per-sample consumption is below all thirteen baselines on both devices. Restricted to the eight baselines measured at the same $256 \times 256$ resolution on GPU, \emph{Colorist} uses $3.8\times$ to $26.8\times$ less energy. The tightest comparison is Contrimix, which runs at $224 \times 224$ on fewer pixels than \emph{Colorist} and still consumes $1.2\times$ more. The two largest GPU factors, $158\times$ for WCT2 and $176\times$ for Modflows, are inflated by resolution, since both operate at $512 \times 512$ and therefore on four times the pixels; they report the cost of those methods as deployed rather than a like-for-like ratio.

Latency and energy therefore point the same way once the device is held fixed: \emph{Colorist} matches the fastest GPU-accelerated baselines while running on CPU, and on the same device it is both faster and lower-energy than every one of them. The energy result is the one that holds everywhere, at every resolution and on both devices, against every method measured.

\begin{table}[htb]
\centering
\caption{Computational efficiency and energy profiling of \emph{Colorist} against the thirteen deep generative baselines of Table~\ref{tab:curated_ssim}, reported on both devices. Res. is the method's own declared native input resolution, which the harness honors and which is therefore not constant across rows; the latency columns give the mean and the $50$th, $90$th, $95$th, and $99$th percentiles per processed image; throughput is images per second; energy is per sample, measured with \texttt{codecarbon}. The GPU block (NVIDIA RTX 6000 Ada, $10{,}000$ iterations) is how the baselines are deployed; the CPU block ($1{,}000$ iterations) is the like-for-like comparison against \emph{Colorist}, which has no GPU implementation and therefore appears once, at the foot of that block. Fifty warmup iterations are discarded before timing in every run, and one image is processed per iteration, so per-sample energy is comparable across rows even though the iteration counts are not. \emph{Colorist} consumes less energy than every baseline on either device.}
\label{tab:runtime_sustainability_supp}
\setlength{\tabcolsep}{3pt}
\resizebox{\textwidth}{!}{%
\begin{tabular}{l|c|rrrrrrr}
\toprule
\multirow{2.5}{*}{Method} & \multirow{2.5}{*}{Res.} & \multicolumn{5}{c}{Latency (ms)} & \multicolumn{1}{c}{\multirow{1.75}{*}{Throughput}} & \multicolumn{1}{c}{\multirow{1.75}{*}{Energy}} \\
\cmidrule(lr){3-7}
 & & \multicolumn{1}{c}{Mean} & \multicolumn{1}{c}{P50} & \multicolumn{1}{c}{P90} & \multicolumn{1}{c}{P95} & \multicolumn{1}{c}{P99} & \multicolumn{1}{c}{(it/s)} & \multicolumn{1}{c}{(mWh/sample)} \\
\midrule
\rowcolor{gray!15} \multicolumn{9}{l}{\textit{GPU (NVIDIA RTX 6000 Ada), $10{,}000$ iterations}} \\
\: AdaIN \venue{ICCV '17} & $256^2$ & 2.07 & 2.07 & 2.09 & 2.10 & 2.14 & 483.52 & 0.2042 \\
\: ArtFlow \venue{CVPR '21} & $256^2$ & 16.87 & 16.87 & 17.13 & 17.22 & 17.48 & 59.27 & 1.4249 \\
\: EFDM \venue{CVPR '22} & $256^2$ & 2.17 & 2.15 & 2.18 & 2.20 & 2.89 & 461.66 & 0.2144 \\
\: IEContrAST \venue{NIPS '21} & $256^2$ & 3.16 & 3.15 & 3.19 & 3.26 & 3.83 & 316.01 & 0.3023 \\
\: MAST \venue{ACM MM '20} & $256^2$ & 3.31 & 3.30 & 3.35 & 3.36 & 3.62 & 301.83 & 0.3093 \\
\: SANET \venue{CVPR '19} & $256^2$ & 3.18 & 3.17 & 3.20 & 3.21 & 3.38 & 314.84 & 0.2964 \\
\: Styleformer \venue{ICCV '21} & $256^2$ & 3.72 & 3.70 & 3.76 & 3.79 & 4.48 & 268.53 & 0.3457 \\
\: StyTr2 \venue{CVPR '22} & $256^2$ & 8.51 & 8.50 & 8.80 & 9.00 & 9.52 & 117.44 & 0.7742 \\
\: Contrimix \venue{COMPAY '24} & $224^2$ & 1.50 & 1.46 & 1.78 & 1.99 & 3.10 & 665.17 & 0.0650 \\
\: SGViTs \venue{MICCAI '24} & $224^2$ & 5.11 & 5.06 & 5.18 & 5.26 & 6.51 & 195.79 & 0.4721 \\
\: StylizingViT \venue{ISBI '26} & $224^2$ & 11.71 & 11.65 & 12.02 & 12.19 & 12.60 & 85.43 & 1.0021 \\
\: Modflows \venue{AAAI '25} & $512^2$ & 109.38 & 109.33 & 109.77 & 109.93 & 110.26 & 9.14 & 9.3640 \\
\: WCT2 \venue{ICCV '19} & $512^2$ & 143.13 & 143.12 & 144.38 & 144.80 & 145.98 & 6.99 & 8.4244 \\
\midrule
\rowcolor{gray!15} \multicolumn{9}{l}{\textit{CPU, $1{,}000$ iterations}} \\
\: AdaIN \venue{ICCV '17} & $256^2$ & 132.20 & 132.95 & 141.10 & 142.56 & 145.00 & 7.56 & 4.2766 \\
\: ArtFlow \venue{CVPR '21} & $256^2$ & 989.50 & 954.39 & 1072.84 & 1076.69 & 1085.22 & 1.01 & 33.1379 \\
\: EFDM \venue{CVPR '22} & $256^2$ & 138.09 & 137.86 & 145.76 & 147.00 & 149.17 & 7.24 & 4.6606 \\
\: IEContrAST \venue{NIPS '21} & $256^2$ & 330.20 & 332.85 & 351.49 & 356.51 & 367.85 & 3.03 & 10.2056 \\
\: MAST \venue{ACM MM '20} & $256^2$ & 449.60 & 451.70 & 482.15 & 493.55 & 515.55 & 2.22 & 13.4629 \\
\: SANET \venue{CVPR '19} & $256^2$ & 297.04 & 299.16 & 316.00 & 320.00 & 327.87 & 3.37 & 9.2891 \\
\: Styleformer \venue{ICCV '21} & $256^2$ & 21158.19 & 21246.01 & 21550.12 & 21601.64 & 21904.99 & 0.05 & 563.6263 \\
\: StyTr2 \venue{CVPR '22} & $256^2$ & 977.97 & 988.17 & 1066.88 & 1088.27 & 1126.09 & 1.02 & 27.5014 \\
\: Contrimix \venue{COMPAY '24} & $224^2$ & 6.22 & 6.14 & 6.51 & 6.82 & 8.20 & 160.78 & 0.2883 \\
\: SGViTs \venue{MICCAI '24} & $224^2$ & 175.02 & 176.40 & 180.88 & 182.91 & 183.86 & 5.71 & 5.7545 \\
\: StylizingViT \venue{ISBI '26} & $224^2$ & 352.37 & 348.91 & 364.31 & 364.59 & 365.37 & 2.84 & 11.7491 \\
\: Modflows \venue{AAAI '25} & $512^2$ & 14373.39 & 14370.37 & 14437.85 & 14489.71 & 14542.44 & 0.07 & 371.6897 \\
\: WCT2 \venue{ICCV '19} & $512^2$ & 8742.16 & 8740.63 & 8833.44 & 8862.15 & 8898.22 & 0.11 & 227.3457 \\
\midrule
\emph{Colorist} (Ours, CPU) & $256^2$ & 2.53 & 2.40 & 2.86 & 3.30 & 4.32 & 395.38 & 0.0532 \\
\bottomrule
\end{tabular}
}
\end{table}


\end{document}